\documentclass[12pt, final]{l4dc2020} 

\usepackage[normalem]{ulem}                         
\usepackage{enumitem}                               
\usepackage{extarrows}                              
\usepackage{verbatim}%

\usepackage{amsmath,amssymb,amsfonts,dsfont} 
\usepackage{bbm}
\usepackage{algorithm,algorithmicx,listings}        
\usepackage[noend]{algpseudocode}             

\usepackage[export]{adjustbox}
\usepackage{makecell,booktabs}
\usepackage[font={small}]{caption}
\usepackage{breakcites}

\def\argmax{\mathop{\arg\max}\limits}

\newcommand{\indicator}{\mathds{1}}

\newcommand{\scaleMathLine}[2][1]{\resizebox{#1\linewidth}{!}{$\displaystyle{#2}$}}
\newcommand{\norm}[1]{\left\lVert#1\right\rVert}
\newcommand{\prl}[1]{\left(#1\right)}
\newcommand{\brl}[1]{\left[#1\right]}
\newcommand{\crl}[1]{\left\{#1\right\}}

\usepackage{mathtools}
\usepackage{pifont}
\usepackage{footnote}
\usepackage{tikz}
\makesavenoteenv{tabular}
\makesavenoteenv{table}
\newcommand{\defeq}{\vcentcolon=}

\def\etal/{et~al.}
\graphicspath{{fig/}}

\newtheorem*{problem*}{Problem}
\newtheorem*{definition*}{Definition}
\newtheorem*{assumption*}{Assumption}
\newtheorem*{proposition*}{Proposition}
\makeatletter
\def\set@curr@file#1{\def\@curr@file{#1}} 
\makeatother

\title[Learning Navigation Costs from Demonstration with Semantic Observations]{Learning Navigation Costs from Demonstration with Semantic Observations}
\usepackage{times}

\author{%
 \Name{Tianyu Wang} \Email{tiw161@eng.ucsd.edu}\\
 \Name{Vikas Dhiman} \Email{vdhiman@eng.ucsd.edu}\\
 \Name{Nikolay Atanasov} \Email{natanasov@eng.ucsd.edu}\\
 \addr Department of Electrical and Computer Engineering, University of California San Diego, La Jolla, CA 92093%
}

\begin{document}

\maketitle

\begin{abstract}
This paper focuses on inverse reinforcement learning (IRL) for autonomous robot navigation using semantic observations. The objective is to infer a cost function that explains demonstrated behavior while relying only on the expert's observations and state-control trajectory. We develop a map encoder, which infers semantic class probabilities from the observation sequence, and a cost encoder, defined as deep neural network over the semantic features. Since the expert cost is not directly observable, the representation parameters can only be optimized by differentiating the error between demonstrated controls and a control policy computed from the cost estimate. The error is optimized using a closed-form subgradient computed only over a subset of promising states via a motion planning algorithm. We show that our approach learns to follow traffic rules in the autonomous driving CARLA simulator by relying on semantic observations of cars, sidewalks and road lanes.
\end{abstract}



\section{Introduction}
\label{sec:introduction}



Autonomous systems operating in unstructured, partially observed, and changing
real-world environments need an understanding of context to evaluate the safety,
utility, and efficiency of their performance. 
For example, while a bipedal robot may navigate along sidewalks, an autonomous car needs 
to follow the road lane structure and the traffic signs. 
Designing a cost function that encodes such rules by hand is cumbersome, 
if not infeasible, especially for complex tasks. 
However, it is often possible to obtain demonstrations of desirable behavior that 
indirectly capture the role of semantic context in the task execution. 
Semantic labels provide rich information about the relationship between object 
entities and their surroundings. 
In this work, we consider an inverse reinforcement learning (IRL) problem in which
observations containing semantic information about the environment are available.


There has been significant progress in semantic segmentation techniques, including deep neural networks for RGB image segmentation~\citep{Papandreou2015WeakSemSeg,Badrinarayanan2017Segnet, Chen2018EncoderSemSeg} or point cloud labeling via a 2D spherical depth projection~\citep{Wu2018SqueezeSeg, Dohan2015LidarSemSeg}. Maps that store semantic information can be generated from segmented images~\citep{Sengupta2012SemMap, Lu2019MonoSemMap}. \citet{Gan2019Bayesian, Sun2018ReccOctoMap} 
generalize binary occupancy grid mapping~\citep{Hornung2013Octomap} to multi-class semantic mapping in 3D. In this work, we parameterize the navigation cost of an autonomous vehicle as a nonlinear function of such semantic features to explain the demonstrations of an expert.

Learning a cost function from demonstration requires a control policy that is differentiable with respect to the cost parameters. 
Computing policy derivatives has been addressed by several sucessful IRL approaches~\citep{Neu2012Apprenticeship, Ratliff2006MMP, Ziebart2008MaxEnt}. Early works assume that the cost is linear in the feature vector and aim at matching the feature expectations of the learned and expert policies. \citet{Ratliff2006MMP} computes subgradients of planning algorithms so that expected reward of an expert policy is better than any other policy by a margin. Value iteration networks (VIN)~\citep{Tamar2016VIN} show that the value iteration algorithm can be approximated by a series of convolution and maxpooling layers, allowing automatic differentiation to learn the cost function end-to-end. \citet{Ziebart2008MaxEnt} develops a dynamic programming algorithm to maximize the likelihood of observed expert data and learns a policy of maximum entropy (MaxEnt) distribution. Many works~\citep{Levine2011Nonlinear, Wulfmeier2016DeepMaxEnt, Song2019IRL} extend MaxEnt to learn a nonlinear cost using Gaussian Processes or deep neural networks. \citet{Finn2016GCL} uses sample-based approximation of the MaxEnt objective on high-dimensional continuous systems. However, the cost in most existing work is learned offline using full observation sequences from the expert demonstrations. A \emph{major contribution} of our work is to develop cost representations and planning algorithms that rely only on causal partial observations.

Achieving safe and robust navigation is directly coupled with the quality of the environment representation and the cost function specifying desirable behaviors. The traditional approach combines geometric mapping of occupancy probability or distance to the nearest obstacle~\citep{Hornung2013Octomap, Oleynikova2017Voxblox} with hand-specified planning cost functions. Recent advances in deep reinforcement learning demonstrated that control inputs may be predicted directly from sensory observations~\citep{visuomotor}. However, special model designs~\citep{Khan2018MACN} that serve as a latent map are needed in navigation tasks where simple reactive policies are not feasible. \citet{Gupta2017CMP} decompose visual navigation into two separate stages explicitly: mapping the environment from first-person RGB images and planning through the constructed map with VIN. Our model also seperates the two stages but integrates semantic information to obtain a richer map representation.
In addition, \citet{Wang2020ICRA} propose a differentiable mapping and planning framework to learn the expert cost function.
They parameterize cost function as a neural network over the binary occupancy map probability, which is integrated from previous partial observations. 
They further propose an efficient A*~\citep{Hart1968Astar} planning
algorithm that computes the policy at the current state and backpropagates
gradients in closed-form to optimize the cost parameters.
We extend their work by incorporating semantic observation to the map representation and evaluating the model in the CARLA autonomous driving simulator~\citep{Dosovitskiy2017CARLA}. 

We propose a model that learns to navigate from first-person semantic observations and make the following contributions. First, we propose a cost function representation composed of a \textit{map encoder}, capturing semantic class probabilities from the streaming observations, and a \textit{cost encoder}, defined as a deep neural network over the semantic features. Second, we optimize the cost parameters using a closed-form subgradient of the cost-to-go only over a subset of promising states, obtained by an efficient planning algorithm. Finally, we verify our model in autonomous navigation experiments in urban environments provided by the CARLA simulator~\citep{Dosovitskiy2017CARLA}.

\begin{figure}[t]
  \centering
  \includegraphics[width=\linewidth,trim=0mm 3mm 0mm 3mm, clip]{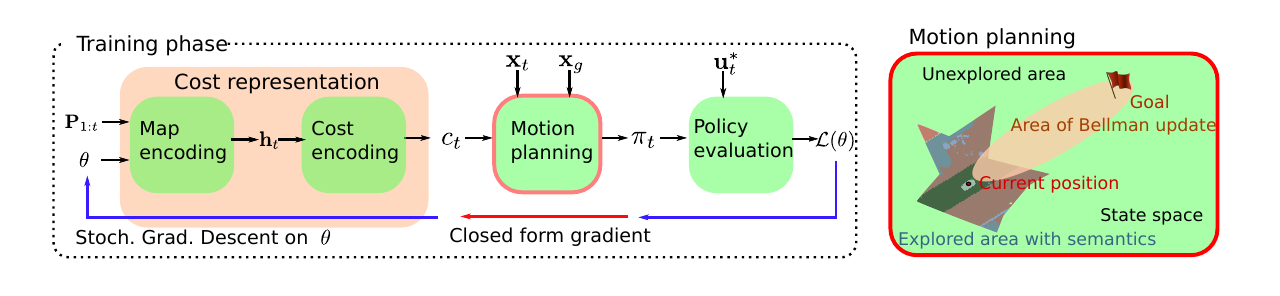}
  \caption{Architecture for cost function learning from demonstrations with semantic observations. 
    Our main contribution is a cost representation, combining a probabilistic \emph{
    semantic map encoder}, with recurrent dependence on semantic observations $\vec{P}_{1:t}$, 
    and a \emph{cost encoder}, defined over the semantic features. Efficient forward policy computation and closed-form subgradient backpropagation are used to optimize the cost representation in order to explain expert behavior.}    
  \label{fig:approach}
\end{figure}

\section{Problem Formulation}
\label{sec:problem_formulation}
Consider a robot navigating in an unknown environment $\mathcal{X}$ with the task of reaching
a goal state $\vec{x}_g \in \mathcal{X}$.
Let $\vec{x}_t \in \mathcal{X}$ be the robot state, capturing its pose, twist,
etc., at discrete time $t$.
For a given control input $\vec{u}_t \in \mathcal{U}$ where $\mathcal{U}$ is 
assumed finite, the robot state evolves according to known deterministic dynamics:
$\vec{x}_{t+1} = f(\vec{x}_t, \vec{u}_t)$.
Let $\mathcal{K} = \crl{0,1,2,\dots,K}$ be a set of class labels, where $k=0$ denotes 
``free'' space and $k \geq 1$ denotes a particular semantic class such as car or tree. 
Let $m^* : \mathcal{X} \rightarrow \mathcal{K}$ be a function specifying the
\textit{true} semantic occupancy of the environment by labeling states with semantic 
classes and $\mathcal{M}$ be the space of possible environment realizations $m^*$.
Let $c^* : \mathcal{X} \times \mathcal{U} \times \mathcal{M} \rightarrow
\mathbb{R}_{\ge 0}$ be a cost function specifying desirable robot behavior in a
given environment, e.g., according to an expert user or an optimal design.
We assume that the robot does not have access to either the true semantic map
$m^*$ or the true cost function $c^*$.
However, the robot is able to obtain point cloud observations 
$\vec{P}_t = \crl{\prl{\vec{p}_l, \vec{y}_l}}_l \in \mathcal{P}$ at each step $t$, 
where $\vec{p}_l \in \mathcal{X}$ is the measurement location and 
$\vec{y}_l$ is an observed semantic likelihood such that 
$\vec{y}_l = \brl{y_l^1, \dots, y_l^K}^T, y_l^k \geq 0, \sum_{k=1}^K y_l^k = 1$, 
whose support is $\mathcal{K} \setminus \crl{0}$. 
In practice, $\vec{y}_l$ can be obtained from a semantic segmentation 
algorithm~\citep{Papandreou2015WeakSemSeg, Badrinarayanan2017Segnet, Chen2018EncoderSemSeg}
that predicts the semantic class of the corresponding measurement location $\vec{p}_l$.
The observed point cloud $\vec{P}_t$ depends on the robot state $\vec{x}_t$ and the environment realization $m^*$. Given a training set 
$\mathcal{D} := \crl{(\vec{x}_{t,n},\vec{u}_{t,n}^*,\vec{P}_{t,n}, 
\vec{x}_{g,n})}_{t=1, n=1}^{T_n, N}$ of $N$ expert trajectories with length 
$T_n$ to demonstrate desirable behavior, our goal is to 
\begin{itemize}[nosep]
    \item learn a cost function estimate $c_t: \mathcal{X} \times \mathcal{U} \times 
    \mathcal{P}^t \times \Theta \rightarrow \mathbb{R}_{\ge 0}$ 
    that depends on an observation sequence $\vec{P}_{1:t}$ from the true latent
    environment and is parameterized by $\vec{\theta} \in \Theta$,
    \item design a stochastic policy $\pi_t$ from $c_t$ such that the robot behavior 
    under $\pi_t$ matches the prior experience $\mathcal{D}$.
\end{itemize}
To balance exploration in partially observable environments with exploitation of promising controls, we specify $\pi_t$ as a Boltzmann policy~\citep{Ramachandran2007BayesianIRL, Neu2012Apprenticeship} associated with the cost $c_t$, 
$\pi_t(\vec{u}_t | \vec{x}_t; \vec{P}_{1:t}, \vec{\theta}) =
\frac{\exp(-Q^*_{t}(\vec{x}_t, \vec{u}_t; \vec{P}_{1:t}, \vec{\theta}))}
{\sum_{\vec{u} \in \mathcal{U}}\exp(-Q^*_{t}(\vec{x}_t, \vec{u}; \vec{P}_{1:t},
\vec{\theta}))}$, 
where the optimal cost-to-go function $Q^*_t$ is:
\begin{align}
\label{eqn:Q}
Q^*_t(\vec{x}_t, \vec{u}_t; \vec{P}_{1:t}, \vec{\theta}) := \min_{\vec{u}_{t+1:T-1}} & \sum_{k=t}^{T-1} c_t(\vec{x}_k, \vec{u}_k ; \vec{P}_{1:t}, \vec{\theta}) \;\;
\text{s.t.}\; \vec{x}_{k+1} = f(\vec{x}_k, \vec{u}_k),\; \vec{x}_T = \vec{x}_g. 
\end{align}%

\begin{problem*}
\label{pb:1}
Given demonstrations $\mathcal{D}$, optimize the cost function parameters $\vec{\theta}$ so that log-likelihood of the demonstrated controls $\vec{u}_{t,n}^*$ is maximized under the robot policies $\pi_{t,n}$:
\begin{equation}
\label{eq:loss}
\min_{\vec{\theta}} \mathcal{L}(\vec{\theta}) \defeq 
- \sum_{n=1}^N \sum_{t=1}^{T_n}\log \pi_{t, n} (\vec{u}_{t,n}^*|\vec{x}_{t,n};
\vec{P}_{1:t, n}, \vec{\theta}).
\end{equation}
\end{problem*}

The problem setup is illustrated in Fig.~\ref{fig:approach}. 
While Eqn.~\eqref{eqn:Q} is a standard deterministic shortest path (DSP) problem, 
the challenge is to make it differentiable with respect to $\vec{\theta}$, 
which is necessary for the loss in~\eqref{eq:loss} to propagate back through 
the DSP problem to update the cost parameters $\vec{\theta}$. 
Once the parameters are optimized, the robot can generalize to navigation tasks 
in new partially observable environments by evaluating the cost $c_t$ based on 
the observations $\vec{P}_{1:t}$ iteratively and (re)computing the associated 
policy $\pi_t$.

\section{Cost Function Representation}
\label{sec:cost_representation}
\newcommand{\tm}{\vec{\psi}}
\newcommand{\tmm}{\tm_l}
\newcommand{\tmmk}{\psi_m^k}
\newcommand{\tmmkp}{\psi_m^{k'}}
\newcommand{\tc}{\vec{\phi}}


We propose a cost function representation with two components: a \emph{semantic map encoder} 
with parameters $\tm$ and a \emph{cost encoder} with parameters $\tc$. 
The model is differentiable by design, allowing its parameters to be optimized
by the subsequent planning algorithm described in Sec.~\ref{sec:cost_learning}. 

\subsection{Semantic Map Encoder}
The semantic probability of different environment areas is encoded in a hidden state $\vec{h}_t$ given the trajectory and observations $\vec{x}_{1:t}, \vec{P}_{1:t}$. 
Specifically, we discretize the state space $\mathcal{X}$ into $J$ cells and let 
$\vec{m} = \brl{m^1, \dots, m^J}^T \in \mathcal{K}^J$ 
be the random vector of true semantic labels over the cells. 
Since $\vec{m}$ is unknown to the robot, we maintain the semantic occupancy 
posterior $\mathbb{P}(\vec{m} = \vec{k} | \vec{x}_{1:t}, \vec{P}_{1:t})$
where $\vec{k} = \brl{k^1, \dots, k^J}^T, k^j \in \mathcal{K}$, given the history 
of states $\vec{x}_{1:t}$ and observations $\vec{P}_{1:t}$. 
The representation complexity may be simplified significantly if one assumes 
independence among the map cells $m^j$: $\mathbb{P}(\vec{m} = \vec{k} | \vec{x}_{1:t}, \vec{P}_{1:t}) = 
\prod_{j=1}^J \mathbb{P}(m^j = k^j | \vec{x}_{1:t}, \vec{P}_{1:t})$.

Inspired by the binary occupancy grid mapping~\citep{Thrun2005PR,Hornung2013Octomap}, 
we extend the recurrent updates for the multi-class semantic probability of 
each cell $m^j$. 
\begin{definition}
The log-odds ratio of semantic classes associated with cell $m^j$ 
at time $t$ is 
\begin{equation}
\label{eq:log_odds}
\vec{h}_{t,j} = \brl{h_{t,j}^0, \dots, h_{t,j}^K}^T, \;\;\;
h_{t,j}^k \defeq \log \frac{\mathbb{P}(m^j = 
k | \vec{x}_{1:t}, \vec{P}_{1:t})}{\mathbb{P}(m^j = 0 | \vec{x}_{1:t}, \vec{P}_{1:t})}
\;\text{for}\; k \in \mathcal{K}.
\end{equation}
\end{definition}
Its recurrent Bayesian update is $h_{t+1,j}^k = h_{t,j}^k 
+ \log\frac{p(\vec{P}_{t+1} | m^j = k, \vec{x}_{t+1})}
{p(\vec{P}_{t+1} | m^j = 0, \vec{x}_{t+1})}$. 
Note that by definition $h_{t,j}^0 = 0$. 
The update increment is a log-odds observation model and we assume the observation $\vec{P}_{t+1}$ given the cell $m^j$ is independent of the previous observations $\vec{P}_{1:t}$. 
The semantic class posterior can be recovered from the semantic log-odds 
ratio $\vec{h}_{t,j}$ via a softmax function 
$\mathbb{P}(m^j = k | \vec{x}_{1:t}, \vec{P}_{1:t}) = 
\sigma^k(\vec{h}_{t,j})$,
where $\sigma : \mathbb{R}^{K+1} \rightarrow \mathbb{R}^{K+1}$ 
satisfies the following properties
\begin{equation}
\label{eqn:softmax}
\sigma(\vec{z}) = \brl{\sigma^0(\vec{z}), \dots, \sigma^K(\vec{z})}^T, \;
\sigma^k(\vec{z}) = \frac{\exp{(z^k)}}{\sum_{k' \in \mathcal{K}} \exp{(z^{k'})}}, \;
\log \frac{\sigma^k(\vec{z})}{\sigma^{k'}(\vec{z})} = z^k - z^{k'}.
\end{equation}

We provide a simple observation model to instantiate Eq.~\eqref{eq:log_odds}.
Consider all cells $m^j$ that lie on the ray between robot state $\vec{x}$ and a labeled
point $\prl{\vec{p}_l, \vec{y}_l}$ in the point cloud $\vec{P}$. 
Let $d(\vec{x}, m^j)$ be the distance between the robot position and the center of mass
of the cell $m^j$.
\begin{definition}
The inverse observation model relating the label of cell $m^j$ to the ray between robot state $\vec{x}$ and labeled point $\prl{\vec{p}_l, \vec{y}_l}$ is defined as a softmax function with parameters $\tmm$,
scaled by the distance, $\delta p_l = d(m^j, \vec{x}) - \norm{\vec{p}_l}_2$, 
which is truncated at a threshold $\epsilon$:
\begin{equation}
\label{eq:simple_inv_obs}
\mathbb{P}(m^j = k | \vec{x}, (\vec{p}_l, \vec{y}_l)) = \begin{cases}
\sigma^k(\textit{diag}(\tmm) \vec{y}_l \delta p_l)
& \text{if } \delta p_l \leq\epsilon\\
\sigma^k(\vec{h}_{0,j})
& \text{if } \delta p_l > \epsilon
\end{cases} \;.
\end{equation}
\end{definition}
The function $\textit{diag}(\cdot)$ returns a diagonal matrix from a vector and the prior occupancy log-odds ratio $\vec{h}_{0,j}$ depends on the environment (e.g., $\vec{h}_{0,j} = \vec{0}$ specifies a uniform prior over the semantic classes).

\begin{proposition}
Given the definitions of the log-odds ratio in Eq.~\eqref{eq:log_odds} and the inverse observation model in Eq.~\eqref{eq:simple_inv_obs}, the log-odds update rule for the semantic probability at cell $m^j$ is
$
\vec{h}_{t+1,j} = \vec{h}_{t,j} 
+ \sum_{\prl{\vec{p}_l, \vec{y}_l} \in \vec{P}_{t+1}} \brl{\vec{g}_j(\vec{x}_t, (\vec{p}_l, \vec{y}_l)) - \vec{h}_{0,j}},
$
where the log-odds inverse observation model for cells $m^j$ along the ray
from $\vec{x}_t$ to $\vec{p}_l$ can be simplied using~\eqref{eqn:softmax} as:
\begin{equation}
\label{eq:simple_inverse_sensor_log_odds}
\vec{g}_j(\vec{x}_t, (\vec{p}_l, \vec{y}_l)) = \begin{cases}
\textit{diag}(\tmm) \vec{y}_l \delta p_l
& \text{if } \delta p_l \leq\epsilon\\
\vec{h}_{0,j} 
& \text{if } \delta p_l > \epsilon
\end{cases} \;.
\end{equation}
\end{proposition}

A more expressive multi-layer neural network may be used to parameterize the inverse observation model instead of the linear transformation $\textit{diag}(\tmm) \vec{y}_l \delta p_l $ of the semantic probability and distance differential along the $l$-th ray in Eq~\eqref{eq:simple_inv_obs}:
\begin{equation}
\label{eq:nn_inverse_sensor_log_odds}
\vec{g}_j(\vec{x}_t, (\vec{p}_l, \vec{y}_l);\tmm) = \begin{cases}
\textbf{NN}(\vec{y}_l, \vec{p}_l, d(\vec{x}_t, m^j) ; \tmm) 
& \text{if } \delta p_l \leq\epsilon\\
\vec{h}_{0,j}
& \text{if } \delta p_l > \epsilon 
\end{cases} \;.
\end{equation}

In summary, the map encoder starts with prior log-odds $\vec{h}_{0}$, 
updates them recurrently via $\vec{h}_{t+1} = 
\vec{h}_{t} + \vec{g}(\vec{x}_t,\vec{P}_t; \tm) - \vec{h}_{0}$,
where the inverse sensor log-odds $\vec{g}_j(\vec{x}_t, (\vec{p}_l, \vec{y}_l) ;\tmm)$ is specified for the $j$-th cell along the $l$-th ray in~\eqref{eq:simple_inverse_sensor_log_odds} or~\eqref{eq:nn_inverse_sensor_log_odds}.
The posterior $\mathbb{P}(\vec{m} = \vec{k} | \vec{x}_{1:t}, \vec{P}_{1:t})$ is the softmax of $\vec{h}_{t}$.


\subsection{Cost Encoder}
The cost encoder uses the semantic occupancy grid prosterior $\sigma(\vec{h}_t)$ to 
define the cost function estimate $c_t(\vec{x},\vec{u})$ at a given state-control 
pair $(\vec{x},\vec{u})$. 
A convolutional neural network (CNN)~\citep{Goodfellow-et-al-2016} with parameters 
$\tc$ can extract cost features from the environment map:
$
c_t(\vec{x},\vec{u}) = \textbf{CNN}(\vec{h}_t, \vec{x}, \vec{u}; \tc).
$
We implement an encoder-decoder neural network architecture~\citep{Badrinarayanan2017Segnet} 
to parameterize the cost function from semantic class probabilities.
The idea is to perform downsamples and upsamples at multiple scales to provide both local and global context between semantic probability and cost.  
\section{Cost Learning via Differentiable Planning}
\label{sec:cost_learning}

We follow the planning algorithm in \citet{Wang2020ICRA} that enables efficient cost optimization and briefly review the steps below. The parameters $\vec{\theta}$ of the cost representation $c_t(\vec{x},\vec{u} ; \vec{P}_{1:t}, \vec{\theta})$ developed in Sec.~\ref{sec:cost_representation} are optimized by differentiating $\mathcal{L}(\vec{\theta})$ in~\eqref{eq:loss} through the DSP problem in~\eqref{eqn:Q}. Motion planning algorithms, such as A*~\citep{ARA}, solve problem~\eqref{eqn:Q} efficiently and determine the optimal cost-to-go $Q_t^*(\vec{x},\vec{u})$ only over a subset of promising states. This is sufficient to obtain the subgradient of $Q_t^*(\vec{x}_t,\vec{u}_t)$ with respect to $c_t$ along the optimal path by applying the subgradient method~\citep{Shor2012Subgradient,Ratliff2006MMP}.

A backwards A* search applied to problem~\eqref{eqn:Q} with start state $\vec{x}_g$, goal state $\vec{x} \in \mathcal{X}$, and predecessors expansions according to transition $f$ provides an upper bound to the optimal cost-to-go: $Q_t^*(\vec{x},\vec{u}) \leq 
c_t(\vec{x},\vec{u}) + g(f(\vec{x},\vec{u}))$, where $g$ are the values computed by A* for expanded nodes in the CLOSED list and visited nodes in the OPEN list. Strict equality is obtained only if $f(\vec{x},\vec{u})$ belongs to the CLOSED list. A Boltzmann policy $\pi_t(\vec{u} \mid \vec{x})$ may be defined using the $g$-values for all $\vec{x} \in \text{CLOSED} \cup \text{OPEN} \subseteq \mathcal{X}$ and a uniform distribution over $\mathcal{U}$ for all other states.

We rewrite $Q_t^*(\vec{x}_t,\vec{u}_t)$ in a form that makes its subgradient with respect to $c_t(\vec{x},\vec{u})$ obvious. Let $\mathcal{T}(\vec{x}_t,\vec{u}_t)$ be the set of feasible trajectories $\vec{\tau}$ of horizon $T$ that start at $\vec{x}_t$, $\vec{u}_t$, satisfy transition $f$ and terminate at $\vec{x}_g$. Let $\vec{\tau}^* \in \mathcal{T}(\vec{x}_t,\vec{u}_t)$ be an optimal trajectory corresponding to the optimal cost-to-go function $Q^*_t(\vec{x}_t,\vec{u}_t)$. Define $\mu_{\vec{\tau}}(\vec{x},\vec{u}) \defeq \sum_{k=t}^{T-1} \indicator_{(\vec{x}_k,\vec{u}_k) = (\vec{x},\vec{u})}$ as a state-control visitation 
function indicating if $(\vec{x},\vec{u})$ is visited by $\vec{\tau}$. The optimal cost-to-go function 
$Q^*_t(\vec{x}_t,\vec{u}_t)$ can be viewed as a minimum over $\mathcal{T}(\vec{x}_t,\vec{u}_t)$ of the inner product of the cost function $c_t$ and the visitation function $\mu_{\vec{\tau}}$:
\begin{equation}
\label{eq:inner_product_q}
Q^*_t(\vec{x}_t,\vec{u}_t) = \min_{\vec{\tau} \in \mathcal{T}(\vec{x}_t,\vec{u}_t)}  
\sum_{\vec{x} \in \mathcal{X},\vec{u}\in\mathcal{U}} c_t(\vec{x},\vec{u}) 
\mu_{\vec{\tau}}(\vec{x},\vec{u})
\end{equation}
where $\mathcal{X}$ can be assumed finite because both $T$ and $\mathcal{U}$ are finite. Applying the subgradient method \citep{Shor2012Subgradient,Ratliff2006MMP} to~\eqref{eq:inner_product_q} shows that $\frac{\partial Q^*_t(\vec{x}_t,\vec{u}_t)}{\partial c_t(\vec{x},\vec{u})} = \mu_{\vec{\tau}^*}(\vec{x},\vec{u})$ is a subgradient of the optimal cost-to-go. This result and the chain rule allow us to obtain a subgradient of $\mathcal{L}(\vec{\theta})$.

\begin{proposition}
\label{prop:chain_rule}
A subgradient of the loss function $\mathcal{L}(\vec{\theta})$ in~\eqref{eq:loss} with respect to $\vec{\theta}$ can be obtained as:
\begin{equation*}
\scaleMathLine{\begin{aligned}
\frac{\partial \mathcal{L}(\vec{\theta})}{\partial \vec{\theta}} &= 
- \sum_{n=1}^N \sum_{t=1}^{T_n} \frac{\partial \log \pi_{t,n}(\vec{u}_{t,n}^* 
\mid \vec{x}_{t,n})}{\partial \vec{\theta}} =- \sum_{n=1}^N \sum_{t=1}^{T_n}\sum_{\vec{u}_{t,n} \in \mathcal{U}} 
\frac{\partial \log \pi_{t,n}(\vec{u}_{t,n}^* \mid \vec{x}_{t,n})}
{\partial Q_{t,n}^*(\vec{x}_{t,n},\vec{u}_{t,n})} 
\frac{\partial Q_{t,n}^*(\vec{x}_{t,n},\vec{u}_{t,n})}
{\partial \vec{\theta}}\\
&=- \sum_{n=1}^N \sum_{t=1}^{T_n}\sum_{\vec{u}_{t,n} \in \mathcal{U}} \prl{\indicator_{\{\vec{u}_{t,n} = \vec{u}_{t,n}^*\}} - 
\pi_{t,n}(\vec{u}_{t,n} | \vec{x}_{t,n})} \!\!\!\!\!\sum_{(\vec{x},\vec{u}) \in \vec{\tau}^*} \!\!\!\!\!
\frac{\partial Q_{t,n}^*(\vec{x}_{t,n},\vec{u}_{t,n})}{\partial c_t(\vec{x},\vec{u})} 
\frac{\partial c_t(\vec{x},\vec{u})}{\partial \vec{\theta}}
\end{aligned}}
\end{equation*}
\end{proposition}

The computation graph implied by Prop.~\ref{prop:chain_rule} is illustrated in Fig.~\ref{fig:approach}. The graph consists of a cost representation layer and a differentiable planning layer, allowing end-to-end minimization of $\mathcal{L}(\vec{\theta})$ via stochastic subgradient descent. Training and testing algorithms are shown in Alg.~\ref{alg:train} and Alg.~\ref{alg:test}.

\begin{figure*}
\begin{minipage}{0.53\linewidth}
\begin{algorithm2e}[H]
\caption{Train cost parameters $\vec{\theta}$}
\label{alg:train}
\SetAlgoLined
\SetKwInOut{Input}{input}\SetKwInOut{Output}{output}
\footnotesize
\Input{$\mathcal{D} \!=\! \crl{(\vec{x}_{t,n},\vec{u}_{t,n}^*,
\vec{P}_{t,n}, \vec{x}_{g,n})}_{t=1, n=1}^{T_n,N}\!\!$}
\While{$\vec{\theta}$ not converged}{
  $\mathcal{L}(\vec{\theta}) \gets 0$\;
  \For{$n = 1, \ldots,N$ \textbf{and} $t = 1,\ldots, T_n$}{
    Update $c_{t,n}$ based on $\vec{x}_{t,n}$ and $\vec{P}_{t,n}$ as in 
    Sec.~\ref{sec:cost_representation}\;
    Obtain $Q_{t,n}^*(\vec{x},\vec{u})$ from~\eqref{eqn:Q} with stage cost $c_{t,n}$\;
    Obtain $\pi_{t,n}(\vec{u} | \vec{x}_{t,n})$ from $Q_{t,n}^*(\vec{x}_{t,n},\vec{u})$\;
    $\mathcal{L}(\vec{\theta}) \gets \mathcal{L}(\vec{\theta}) -\log \pi_{t, n} 
    (\vec{u}_{t,n}^*|\vec{x}_{t,n})$\;
  }
  Update $\vec{\theta} \gets \vec{\theta} - \alpha \nabla \mathcal{L}(\vec{\theta})$ via Prop.~\ref{prop:chain_rule}\;
}
\end{algorithm2e}
\end{minipage}%
\hfill%
\begin{minipage}{0.46\linewidth}
\begin{algorithm2e}[H]
\caption{Test control policy $\pi_t$}
\label{alg:test}
\SetAlgoLined
\SetKwInOut{Input}{input}\SetKwInOut{Output}{output}
\footnotesize
\Input{Start state $\vec{x}_s$, goal state $\vec{x}_g$, cost parameters $\vec{\theta}$}
\texttt{\\}
Current state $\vec{x}_t \gets \vec{x}_s$\;
\While{$\vec{x}_t \neq \vec{x}_g$}{
  Make an observation $\vec{P}_t$\;
  Update $c_t$ based on $\vec{x}_t$ and $\vec{P}_t$ as in Sec.~\ref{sec:cost_representation}\;
  Obtain $\pi_t(\vec{u} | \vec{x}_t)$ from $Q_t^*(\vec{x}_t,\vec{u})$ \;
  Update $\vec{x}_t \gets f(\vec{x}_t, \vec{u}_t)$ via $\vec{u}_t := 
  \argmax_{\vec{u}} \pi_t(\vec{u}|\vec{x}_t)$\;
}
\end{algorithm2e}
\end{minipage}
\end{figure*}

\section{Experiments}
\label{sec:experiments}

\subsection{Experiment Setup}
We evaluate our approach using the CARLA simulator (0.9.6)~\citep{Dosovitskiy2017CARLA},
which provides high-fidelity autonomous vehicle simulation in urban environments.
Demonstration data for training the cost function representation is collected from 
maps $\left\{Town01, Town02, Town03,\right.$ $\left.Town04\right\}$, while map $Town05$ is used for testing. 
$Town05$ is the largest map and includes different street layouts, junctions, 
and a freeway. 
In each map, we collect $100$ expert trajectories by running the autonomous navigation 
agent provided by the CARLA Python API. 
The expert finds the shortest path between two query points, while respecting 
traffic rules, such as staying on the road, and keeping in the current lane. 
Features not related to the experiment are disabled, including spawning 
other vehicles and pedestrians, and changing traffic signal. 
Each vehicle trajectory is discretized into a $128\times128$ grid of $1$ meter resolution.
The robot state $\vec{x}$ is the grid cell location while the control $\vec{u}$ takes the 
robot to one of its $8$ neighbor grid cells.
Trajectories that do not fit in the grid are discarded.

The ego vehicle is equipped with a lidar sensor that has $20$ meters 
maximum range and $360^{\circ}$ horizontal field of view.
The vertical field of view ranges from $0^{\circ}$ 
(facing forward) to $-30^{\circ}$ (facing down) with $5^{\circ}$ resolution. 
A total of $56000$ lidar rays is generated per scan $\vec{P}_t$
while each point measurement is returned only if it hits an obstacle.
The ego vehicle is also equipped with $4$ semantic segmentation cameras 
that display objects of 13 different classes in RGB images, 
including road, road line, sidewalk, vegetation, car, building, etc.
The $4$ cameras face front, left, right and rear, each capturing a $90^{\circ}$ 
horizontal field of view. 
The semantic label of each lidar point is retrieved from the semantic segmentation
image by projecting the lidar point in the camera's frame.

\subsection{Models and Metrics}
We compare our model with two baseline algorithms: \citet{Wulfmeier2016DeepMaxEnt} 
and \citet{Wulfmeier2016DeepMaxEnt}~+~semantics.
\citet{Wulfmeier2016DeepMaxEnt} use a neural network to learn a cost from
lidar point clouds without semantics.
The input to the neural network is a grid that stores the mean and variance of 
points in each cell, as well as a binary indicator of cell visibility. 
We augment the grid features with the mode of semantic labels in each cell 
to get the model \citet{Wulfmeier2016DeepMaxEnt}~+~semantics as a fair comparison with ours.
Neural networks are implemented in the PyTorch library~\citep{Paszke2019Pytorch} and
trained with the Adam optimizer~\citep{Kingma2014ADAM} until convergence.

The evaluation metrics include: 
negative log-likelihood (\textit{NLL}), control accuracy (\textit{Acc}),
trajectory success rate (\textit{Traj. Succ. Rate}) and Modified Hausdorff Distance 
(\textit{MHD}). 
More precisely, given a test set 
$\mathcal{D}_{test} = \crl{(\vec{x}_{t,n},\vec{u}_{t,n}^*,\vec{P}_{t,n}, 
\vec{x}_{g,n})}_{t=1, n=1}^{T_n, N}$ and a learned policy $\pi$ with paramters $\vec{\theta}^*$, we define $\textit{NLL}(\mathcal{D}_{test}, \pi) = 
- \frac{1}{\sum_{n=1}^{N} T_n} \sum_{n=1, t=1}^{N, T_n} 
\log \pi_{t,n}(\vec{u}_{t,n}^* | \vec{x}_{t,n}; \vec{P}_{1:t,n}, \vec{\theta}^*) $
and 
$\textit{Acc}(\mathcal{D}_{test}, \pi) = \frac{1}{\sum_{n=1}^{N} T_n} \sum_{n=1, t=1}^{N, T_n}
\indicator_{\crl{\vec{u}_{t,n}^* = \argmax \pi_{t,n}(\cdot | \vec{x}_{t,n}; 
\vec{P}_{1:t,n}, \vec{\theta}^*)}}$.
\textit{Traj. Succ. Rate} records the success rate of the learned policy by 
iteratively rolling out its predicted controls.
A trajectory is regarded as successful if it reaches the goal within twice the number of 
steps of the expert trajectory without hitting an obstacle.
\textit{MHD} compares the rolled out trajectory $\vec{\tau}_{L}$ by the learned policy 
and the expert trajectory $\vec{\tau}_E$ and is defined as:
$
\textit{MHD} (\vec{\tau}_{L}, \vec{\tau}_{E}) = \max{
\crl{ \frac{1}{T_L}\sum_{t=1}^{T_L} d(\vec{\tau}_{L}^t, \vec{\tau}_{E}),
\frac{1}{T_E}\sum_{t=1}^{T_E} d(\vec{\tau}_{E}^t, \vec{\tau}_{L}) }}
$
where $d(\vec{\tau}_{A}^t, \vec{\tau}_{B})$ measures the minimum Euclidean distance from
state $\vec{\tau}_{A}^t$ to any state in $\vec{\tau}_{B}$.

\subsection{Results and Discussion}
\begin{figure*}[t]
\begin{tabular}{c c c c c}
    \toprule
    Model & \makecell{\textit{NLL}} & \makecell{\textit{Acc} (\%)} & \makecell{\textit{Traj. Succ. Rate} (\%)} & \makecell{\textit{MHD}} \\
    \midrule
    ~\cite{Wulfmeier2016DeepMaxEnt} & 0.595 & 86.1 & \textbf{92} & 4.521 \\
    \makecell{~\cite{Wulfmeier2016DeepMaxEnt} $+$ semantics} & 0.613 & 82.7 & 88 & 4.479 \\
    Ours & \textbf{0.446} & \textbf{90.5} & 91 & \textbf{3.036} \\
    \bottomrule
\end{tabular}
\caption{Test result from CARLA \textit{Town05} map. Best model for each evaluation metric is in bold.}
\label{fig:test_results}
\end{figure*}
\begin{figure*}[t]
    \centering
    \includegraphics[width=\linewidth]{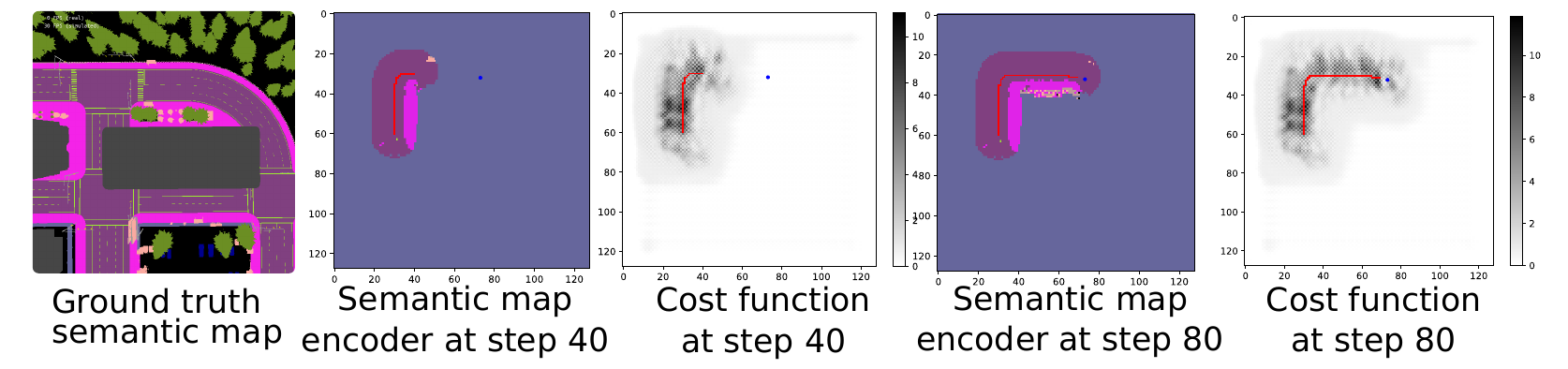}
    \caption{
    Example of a predicted trajectory in red at an intersection and the goal in blue.
    The groud truth semantic map, predicted semantic map and cost map at two time steps are shown. 
    Our model learns the sidewalk is costly to traverse.}
    \label{fig:map_cost}
\end{figure*}

Fig.~\ref{fig:test_results} shows the performance of our model versus~\citet{Wulfmeier2016DeepMaxEnt} and~\citet{Wulfmeier2016DeepMaxEnt} + semantics
using the metrics described above. 
Ours learns to generate policies closest to the expert in new environments by
scoring best in \textit{NLL} and \textit{Acc}. 
The predicted trajectory is also closest to the expert by achieving the minimum \textit{MHD}.
The results demonstrate that the semantic map encoder captures more geometric as well as 
semantic information so that the cost function can be optimized and generate trajectories 
which match the expert behaviors.
We notice that simply taking the mode of the semantic labels in each grid cell degrades the 
performance of \citet{Wulfmeier2016DeepMaxEnt}.
We conjecture that taking the mode is a deterministic assignment, which could 
provide conflicting semantic information, while our model endorses a probabilistic semantic 
map encoder with Bayesian updates to avoid information loss.
Fig.~\ref{fig:map_cost} shows an example of the predicted trajectory at an intersection.
The semantic map visualizes the class of highest probability,
which mostly reflects the ground truth.
Sub-cell objects like roadlines are captured in the semantic map distribution but not visualized in the most probable class.
It is interesting to find that our model assigns low cost to road in front of the robot, medium cost for sidewalks, and high cost to road behind itself. 
This cost assignment is actually effective for the robot to navigate to the goal.  
\section{Conclusion}
\label{sec:conclusion}

We propose an inverse reinforcement learning approach for infering navigation costs from demonstrations with semantic observations. 
Our model introduces a new cost representation composed of a probabilistic semantic occupancy encoder and a cost encoder defined over the semantic features. 
The cost function can be optimized via backpropagation with closed-form (sub)gradient.
Experiments in the CARLA simulator show that our model outperforms methods that
do not encode semantic information probabilistically over time.
Our work offers a promising solution for learning semantic features in navigation and may enable efficient online learning in challenging conditions.

\acks{We gratefully acknowledge support from NSF CRII IIS-1755568, ARL DCIST CRA W911NF-17-2-0181, and ONR SAI N00014-18-1-2828.}

\newpage
{\small
\bibliography{bib/ref}
}

\end{document}